\documentclass[a4paper,5pt,intlimits,twopages]{iopart}

\expandafter\let\csname equation*\endcsname\relax
\expandafter\let\csname endequation*\endcsname\relax

\usepackage[margin=2cm]{geometry}
\usepackage[utf8]{inputenc}
\usepackage[tbtags]{amsmath}
\usepackage{amsthm,amssymb,dsfont,mathrsfs,bbm,xfrac,tabularx,booktabs}
\usepackage{comment}
\usepackage{mathtools}
\usepackage{stmaryrd}
\usepackage{scalerel}
\usepackage{xcolor}
\usepackage{subcaption}
\usepackage{xfrac}

\usepackage{hyperref}
\usepackage[capitalise]{cleveref}
\usepackage{tikz}
\usepackage{lmodern}
\usepackage[T1]{fontenc}
\usepackage{natbib}

\allowdisplaybreaks

\begin{document}

\title{Theory of Speciation Transitions in Diffusion Models with General Class Structure}

\author{Beatrice Achilli$^{1}$, Marco Benedetti$^{1}$, Giulio Biroli$^{2}$ and Marc M\'ezard$^{1}$}

\address{$^{1}$Department of Computing Sciences, Bocconi University, Milan, Italy}
\address{$^{2}$Laboratoire de Physique de l’\'Ecole Normale Sup\'erieure, ENS, Universit\'e PSL, CNRS, Sorbonne Universit\'e, Universit\'e de Paris, F-75005 Paris, France.}

\begin{abstract}

Diffusion Models generate data by reversing a stochastic diffusion process, progressively transforming noise into structured samples drawn from a target distribution. Recent theoretical work has shown that this backward dynamics can undergo sharp qualitative transitions, known as speciation transitions, during which trajectories become dynamically committed to data classes. Existing theoretical analyses, however, are limited to settings where classes are identifiable through first moments, such as mixtures of Gaussians with well-separated means.
In this work, we develop a general theory of speciation in diffusion models that applies to arbitrary target distributions admitting well-defined classes. We formalize the notion of class structure through Bayes classification and characterize speciation times in terms of free-entropy difference between classes. This criterion recovers known results in previously studied Gaussian-mixture models, while extending to situations in which classes are not distinguishable by first moments and may instead differ through higher-order or collective features. Our framework also accommodates multiple classes and predicts the existence of successive speciation times associated with increasingly fine-grained class commitment.
We illustrate the theory on two analytically tractable examples: mixtures of one-dimensional Ising models at different temperatures and mixtures of zero-mean Gaussians with distinct covariance structures. In the Ising case, we obtain explicit expressions for speciation times by mapping the problem onto a random-field Ising model and solving it via the replica method. Our results provide a unified and broadly applicable description of speciation transitions in diffusion-based generative models.
\end{abstract}

\section{Introduction}
Since their introduction in the Machine Learning framework, Diffusion Models (DMs) \citep{sohldickstein2015deep, ho2020denoising} have established themselves as the state-of-the-art tool for image and video generation \citep{cuiAnalysisLearningFlowbased2023,chenSamplingEasyLearning2023,albergoStochasticInterpolantsUnifying2025}. 
Diffusion Models sample their target probability distribution through a denoising procedure. Assume that data are $N$-dimensional variables distributed according to a probability law $P(a)$. Samples from $P(a)$ can be obtained by initializing a $N$-dimensional vector to iid standard Gaussian entries, and follow the so called \textit{backward diffusion process}, defined by the  stochastic differential equation
\begin{equation}
    -dy_t = y_t dt + 2 \mathcal{S}(y_t, t) dt + \sqrt{2} d\tilde{W}_t.
\end{equation}
where $d\tilde W_t$ is a Wiener process, and the drift term $\mathcal{S}(y_t, t)$, called score function, is what guarantees that at $t=0$ we sample from the target distribution $P(a)$. Its exact expression is $\mathcal{S}(x_t, t) = \nabla_x\log  P_t(x_t)$, where 
\begin{equation}
    P_t(x_t)=\int da P(a)\frac{1}{(2\pi\Delta_t)^{\frac{N}{2}}}\exp \left(-\frac{(x-ae^{-t})^2}{2\Delta_t}\right)
\end{equation}
is the probability distribution of a noise corrupted sample, obtained by sampling $a\sim P(a)$ at $t=0$ and following until time $t$ the \textit{forward diffusion} Ornstein-Uhlenbeck process
\begin{equation}
    dx_t = -x_t dt + \sqrt{2} dW_t.
\end{equation} 

A recent research line aims to characterize theoretically sharp transitions in the qualitative behavior of trajectories during the backward process. Such Symmetry Breaking events \citep{raya2023spontaneous, biroli2024dynamical, ambrogioniStatisticalThermodynamicsGenerative2025, sclocchiPhaseTransitionDiffusion2025, behjooUTurnDiffusion2025} were named speciation transitions. They can be illustrated by the following example. Consider an image dataset comprising photos of cats, dogs, eagles and seagulls. At any instant of the backward process, one can try to guess which of the four types of animals will be represented in the final image, at $t=0$. At the beginning of the backward process, the image is just Gaussian noise, and it is impossible to attribute it to any of the four categories. As backward diffusion proceeds, features from the target distribution gradually emerge, and one is able to guess the final image better than random. 
Over time, the prediction for a given single trajectory will change, even back and forth, between classes, but well established time windows exist, during which such fluctuations happen only between a subset of the classes. In our example, there will be a time window during which the prediction fluctuates between cat and dog, but never to a bird, or conversely, between seagull and eagle, but never to a quadruped. 
The boundaries of such time windows are called \textit{speciation times}. After each speciation, backward trajectories become committed to a smaller subset of classes in the dataset. Conversely, during the forward diffusion process, these transitions mark moments when it becomes hard to guess which class generated the image, now corrupted by increasing amounts of noise. 

The timescale of the first speciation transition, separating a first regime, where backward trajectories are purely noisy, and a second one, when features of the data distribution start to emerge, is computed in \cite{biroli2024dynamical}, where a spectral criterion is derived. For data coming from a mixture of two Gaussians with different means (separated by a distance of the order $\sqrt{N})$, they have obtained a speciation time $t_S=\frac{1}{2}\log N$. 
The extension to more than two Gaussians, and multiple speciation times was discussed in \cite{pavasovic2025classifier}.
The methods adopted in these works have one main limitation: they apply only to target probability distributions where data comes from spatially well separated classes identifiable by the first moments of the associated probability distributions. 

In this work we extend the treatment of speciation to a more general setting which only requires that there exist well-defined classes, i.e. that one can assign a typical configuration to a given class almost surely. In order to do that, we give a precise definition, based on the concept of Bayes classifiers, for decomposing a target probability distribution into classes. 
Then, we introduce a criterion to identify speciation times based on the error probability of a Bayes classifier used to infer the component of origin of a forward diffused data point. This general criterion recovers the previous theory, but it also allows to study more general cases in which the existence of well-defined classes is not identifiable by first moments. Moreover, within this general framework, we also analyze the case of multiple classes leading to several speciation times. 

To illustrate the method, we will use as prototypical models first a mixture of 1D Ising models with different temperatures and then a mixture of Gaussians with zero means and different covariances. Interestingly, we are able to obtain analytical expressions for the speciation times in the 1D Ising mixture by mapping the problem into a 1D Ising model with random field, following the replica calculation in \citep{weigt1996replica, lucibello2014onedimensional}. 

\section{Bayes attribution and pure densities mixtures}\label{sec:densities-mixture}
We are interested in studying probability distributions whose samples can be labeled as belonging to one among a number $R$ of ``well distinguished'' classes, or components. To model this, we will assume that 
\begin{equation}\label{eq:mixture}
    P(a)=\sum_{r=1}^R w_r P_r(a),
\end{equation}
where $w_r$ are non-negative weights with $\sum_r w_r=1$, and $P_r(a)$ represent the different components. Each sample will be assigned to a class $s$ on the basis of the \textit{Bayesian attribution to component}: given a sample $a$ drawn from $P(a)$, we compute 
\begin{equation}
    P(s \mid a)=\frac{P(s,a)}{P(a)}=\frac{w_s P_s(a)}{P(a)},
    \label{eq:bayest0}
\end{equation}
and we assign it to class $\mathrm{argmax}\,P(s \mid a)$. Given $P(a)$, there are many ways to decompose it in classes. To address this ambiguity, we shall restrict our analysis to \textit{Proper Density Decompositions}, requiring that, as $N$ increases, the Bayesian Classifier is able to attribute a typical noise-corrupted sample from $P(a)$ with certainty to one of the components of $P(a)$, for any small but finite amount of noise. Formally, let $\tilde{a}_r= a_r+\eta \,\mathcal{N}(0,\mathbb{I})$ be the random variable obtained by adding i.i.d. Gaussian noise to each feature of $a_r$,  sampled from $P_r(a_r)$. We will say that Eq.~\eqref{eq:mixture} forms a \textit{Proper Density Decomposition} of $P(a)$ if, for any $r$ and $s\neq r$, there exists $\eta=O_N(1)$ and $\epsilon(N)=o_N(1)$ such that $P(s \mid \tilde{a}_r)\leq\epsilon(N)$ with high probability (w.h.p.) over the statistics of $\tilde{a}_r$. 

The property above is a form of concentration when $N\rightarrow \infty$. In the following, we assume that it results from a large deviation form of the probability distribution which is the one that naturally emerges in statistical physics and, also, several high-dimensional problems: assuming $a_r \sim P_r$, then the numbers $P_s(a_r)$, $s=1,\ldots,n$ are random variables, and they can be expressed as $P_s(a)=e^{Nf_{s}(a,N)}$, where $f_{s}(a,N)= (1/N) \log P_s(a)$. If $Nf_{s}(a_s,N) \gg Nf_{r}(a_s,N)$ w.h.p. over the sampling of $a_s$ for any $r\neq s$ as $N\to\infty$, we have a \textit{Proper Density Decomposition}. If Eq.~\eqref{eq:mixture} is a \textit{Proper Density Decomposition} and the distribution of $f_{s}(a_r,N)$ concentrates at large $N$ towards its mean, will say that  forms a \textit{Pure Density Decomposition}. In that case, one can write
\begin{equation}
    \label{eq:P_from_F_t0}
    P_s(a_r)=e^{Nf_{rs}+o(N)}, \qquad f_{rs}=\frac{1}{N}\langle \log  P_s(a_r) \rangle_{a_r},
\end{equation}
where the average is taken over the distribution of diffused samples that originate from component $r$, and the $o(N)$ contribution encapsulates the dependence on the specific realization of the disorder $a_r$. This self-averaging property of $f_{s}(a_r,N)$ does not hold, for example, when any of the $P_r(a)$ can itself be properly decomposed. In this sense, self-averaging is a signature of a ``minimal'' proper decomposition of the measure. The concept of pure densities has also been introduced in \cite{biroli2024kernel}.

\section{General criterion for speciation}\label{sec:general-criterion}
Given a Proper Density Decomposition $P(a)=\sum_{r=1}^R w_r P_r(a)$, as defined in Sec.~\ref{sec:densities-mixture}, we need a rigorous notion of attribution to a component of a forward-diffused data point $x$. The most natural way is generalizing \textit{Bayesian attribution to component} to the noise corrupted sample, namely we compute
\begin{equation}
\label{eq:bayestfinite}
    P(s \mid x;t)=\frac{P(s,x;t)}{P(x;t)}=\frac{w_s P_s(x;t)}{P(x;t)},
\end{equation}
where $P(x;t)=\sum_s w_s P_s(x;t)$ and 
\begin{equation}
    P_s(x;t)=\int d^Na \ P_s(a) \exp\left(-\frac{(x-a e^{-t})^2}{2 \Delta_t}\right),
\end{equation}
and we assign it to class $\mathrm{argmax}\,P(s \mid x;t)$. 
We say that $r$ and $s\neq r$ are well distinguishable classes at time $t$ if $\epsilon(N)=o_N(1)$ exists, such that $P(s \mid x;t)\leq\epsilon(N)$ w.h.p. over the statistics of $x$, obtained by sampling $a$ from $P_r(a)$ and diffusing forward for time $t$. Otherwise, $P_r(x;t)$ and $P_s(x;t)$ are both finite w.h.p., and we say that class $r$ is merged with class $s$ at time $t$.
When $x$ is obtained by diffusing a sample originated from component $r$, each of the $P_s(x;t)$ is reminiscent of the partition function of a disordered system. Disorder is represented by the external field $x$. It is then natural to write 
\begin{equation}
    \label{eq:P_from_F}
    P_s(x;t)=e^{Nf_{rs}(t)+\delta f_{rs}(x, t)}, \qquad f_{rs}(t)=\frac{1}{N}\langle \log  P_s(x;t) \rangle_r
\end{equation}
where the average is taken over the distribution of diffused samples that originate from component $r$, and the $f_{rs}(x, t)=o(N)$ contribution encapsulates the dependence on the specific realization of the disorder $x$. Component $r$ can be reliably identified by the Bayesian Classifier as the origin of the trajectory as long as 
\begin{equation}
\label{eq:classification_condition}
Nf_{ss}(t)+\delta f_{ss}(x, t) \gg Nf_{rs}(t)+\delta f_{rs}(x, t) \qquad \forall s\neq r
\end{equation}
w.h.p. over the statistics of $x$. When this condition is not met, component $r$ is merged at time $t$ with component $s$. Hence, two things happen at once: the Bayes Classifier starts assigning finite probability to more than one class, and with finite probability $\mathrm{argmax}\,P(s \mid x;t)$ misattributes the origin of the trajectory. From \cref{eq:classification_condition}, one can see that merging occurs when the difference between average free energies becomes comparable with their fluctuations, marking the speciation time $t_{rs}$.
\begin{align}
\label{eq:speciation_crit}
|f_{rr}(t_{rs})-f_{rs}(t_{rs})| = K\cdot \sqrt{\text{Var}\left[\frac{1}{N}\log  P_r(x;t_{rs})-\frac{1}{N}\log  P_s(x;t_{rs})\right]}.
\end{align}
The arbitrary constant $K$ is related to how stringent we are in requiring that no misattributions happen between components that are not yet merged. Notice that this criterion involves only the two components whose speciation time we are predicting, regardless of how many components coexist in the target probability distribution. The presence of other components is irrelevant as far as speciation is concerned. As we will see, on the timescale of merging, the difference between the average free energies becomes $O_N(1)$ w.h.p..

\subsection{Large N behavior of speciation time}\label{sec:large-time}
Starting from Eq.~\eqref{eq:speciation_crit} one can derive the expected scaling for the speciation time.
As we will see, speciation times diverge as $\log{N}$ in the large $N$ limit. This allows to simplify the computation of both sides in \cref{eq:speciation_crit}, in terms of an expansion in $e^{-2t}$. 
Explicitly, the average free entropy difference reads 
\begin{equation}
	f_{rr}(t)-f_{rs}(t) = \frac{1}{N}\left[\int dx P_{r}(x;t) \log  P_{r}(x;t)-\int dx P_{r}(x;t) \log  P_{s}(x;t)\right].
\end{equation}
Notice that this can be seen as the Kullback-Leibler divergence between the components $P_r$ and $P_s$ of the mixture.
At large forward times one can approximate $P_{r}(x;t)$ by expanding in $e^{-2t}$ its exact expression. The result is a Gaussian distribution (see e.g. \cite{biroli2024dynamical}) with mean $\mu_s$ and variance $\Sigma_s$. Then, the asymptotic average free entropy difference is a Kullback-Leibler divergence between Gaussians 
\begin{align}
\label{eq:asympt_diff}
	D_{\mathrm{KL}}(\mathcal N(\mu_r,\Sigma_r)\,\|\,\mathcal N(\mu_s,\Sigma_s))=\frac{a_{rs}}{2} (e^{-2t} + e^{-4t}) + \frac{C_{rs}}{4} e^{-4t} + e^{-4t} S_{rs}
\end{align}
The values of $\mu_r,\,\Sigma_r,\, a_{rs}$ and $C_{rs}$ are simple functions of the first and second moments of the $P_s(x;t=0)$ distributions. In particular, $a_{rs}$ and $S_{rs}$ are zero if the components have the same mean (see \ref{app:large-time} for details). Leveraging the Gaussian approximation for $P_r(x;t)$ at large $t$, one can also approximate the rhs of \cref{eq:speciation_crit}:
\begin{equation}
\label{eq:asympt_var}
\mathrm{Var}\left[\frac{1}{N}\log P_r(x;t) - \frac{1}{N}\log P_s(x;t)\right] = \frac{a_{rs}}{N} (e^{-2t} + 2e^{-4t}) + \frac{C_{rs}}{2N}e^{-4t}+o(e^{-4t}).
\end{equation}
In this large $t$ regime, leveraging  \cref{eq:asympt_diff,eq:asympt_var,eq:speciation_crit} the criterion for speciation reads
\begin{enumerate}
    \item If $a_{rs} \ne 0$, i.e. there is separation of first moments, 
	\begin{equation}
    \label{eq:asympt_spec_a}
        \frac{a_{rs}}{2} e^{-2t} = K\cdot \sqrt{\frac{a_{rs}}{N}} e^{-t} \implies t_{rs}= \frac12 \log N -\frac{1}{2}\log \left(\frac{4}{a_{rs}}\right) -  \log K.
	\end{equation}
    This recovers the speciation time scaling obtained in \cite{biroli2024dynamical}.
    \item If $a_{rs} = 0$,
	\begin{equation}
    \label{eq:asympt_spec_c}
		\frac{C_{rs}}{4}e^{-4t} = K\cdot \sqrt{\frac{C_{rs}}{2 N}} e^{-2t}  \implies t_{rs}=\frac{1}{4}\log N -\frac{1}{4}\log\left(\frac{8}{C_{rs}}\right) - \frac{\log K}{2}.
	\end{equation}
    This result extends previous results on speciation time to the cases where the class distribution do not have any first moment.
\end{enumerate}

\begin{figure}[t!]
  \centering
        \includegraphics[width=0.3\textwidth]{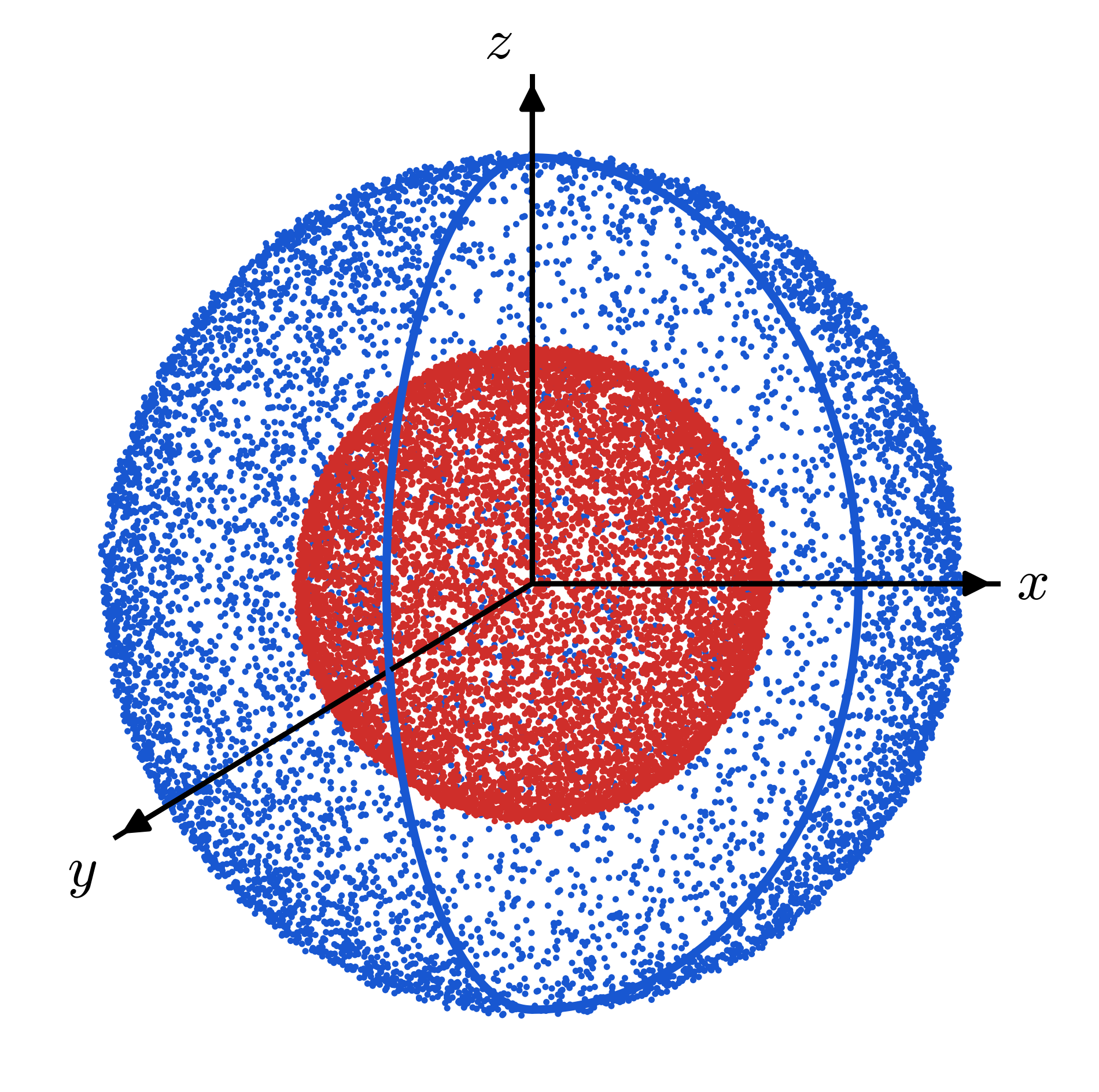}
      \caption{Points scattered on two concentric sphere offer an example of a Pure Density Decomposition, where the two components cannot be distinguished on the basis of their mean. }
    \label{fig:distribution_example}
\end{figure}
\noindent{
\Cref{fig:distribution_example} offers an example of a Pure Density Decomposition, where the two components cannot be distinguished on the basis of their mean. \cref{eq:asympt_spec_a} and \cref{eq:asympt_spec_c} show that the arbitrary constant $K$ performs a shift in the speciation time estimates, irrespective of the target measure. Notice that the leading behavior of the speciation times at large $N$ is independent on the details of the target measure. On the other hand, the finite corrections separating speciation times within components of a given target distribution depend on the specific target distribution, and on the specific components under consideration, through $a_{rs} $ and $C_{rs}$. In both cases, on the timescale of speciations the free entropy differences are $f_{rr}(t_{rs})-f_{rs}(t_{rs}) = O(1/N)$. }

\section{Two simple examples: high dimensional Gaussian data}\label{sec:toy-models}
In this section we illustrate our speciation time predictions on two simple target probability distributions.
For this examples, we can derive speciation time independently of our criterion by analyzing the explicit transition in the shape of the diffusive potential.
We first present a case with first moments separation, namely a mixture of two Gaussians with separate means, which has been analyzed in \cite{biroli2024dynamical}.
Then we proceed to a case without first moments separation, namely again a mixture of two Gaussians centered in zero, with different variances.
In both settings, the speciation times obtained from the potential argument agree with the scalings reported in Sec.~\ref{sec:large-time}.

\subsection{Gaussian Mixture with different means}\label{sec:mixture-mean}
Consider a balanced mixture of two multivariate Gaussians with means $\pm m$ and the same isotropic variance $\sigma^2$
\begin{equation}
    P_t(x) = \frac{1}{2} \frac{1}{(2\pi \Gamma_t)^{N/2}} e^{-\frac{\|x-me^{-t}\|^2}{2\Gamma_t}} + \frac{1}{2} \frac{1}{(2\pi \Gamma_t)^{N/2}} e^{-\frac{\|x+me^{-t}\|^2}{2\Gamma_t}},
\end{equation}
where $\Gamma_t=\sigma^{2}e^{-2t}+(1-e^{-2t})$, and assume $\|m\|^2 = N \tilde{\mu}^2$ so that in large dimensions they are well separated. This setting enters the case $a_{rs}\neq 0$, for which our criterion predicts a speciation time $t\sim 1/2 \log{N}$. We can verify this by looking at the diffusive potential. This  was previously done in \cite{biroli2024dynamical}, we include it here for completeness.
The score function reads
\begin{equation}
    S_t(x) = -\frac{x}{\Gamma_t} + m \frac{e^-t}{\Gamma_t} \tanh{\left(x\cdot m \frac{e^-t}{\Gamma_t}\right)}.
\end{equation}
The reverse-time backward diffusion process for $x_t \in \mathbb{R}^N$ is the Ornstein-Uhlenbeck process:
\begin{equation}
\label{eq:score_guided_diffusion}
    dx = \left(x + 2 S_t(x) \right) dt + \sqrt{2} dW_t
\end{equation}
where $dW_t$ is standard Brownian motion. 
Introducing the overlap $q(t) = \frac{1}{\sqrt{N}}m\cdot x_t$, we can obtain a closed backward stochastic equation, defined by the potential
\begin{equation}
    V(q,t)= \frac{1}{2}q^2 - 2 \tilde{\mu}^2 \log \cosh \left( q e^{-t} \sqrt{N}\right)
\end{equation}
This potential shows a transition: it is quadratic for large times, then at $t = \frac{1}{2}\log N $ it develops a double well structure.

\subsection{Gaussian Mixture with different variances}\label{sec:mixture-variance}
\begin{figure}[t]
    \centering
    \includegraphics[width=0.45\textwidth]{"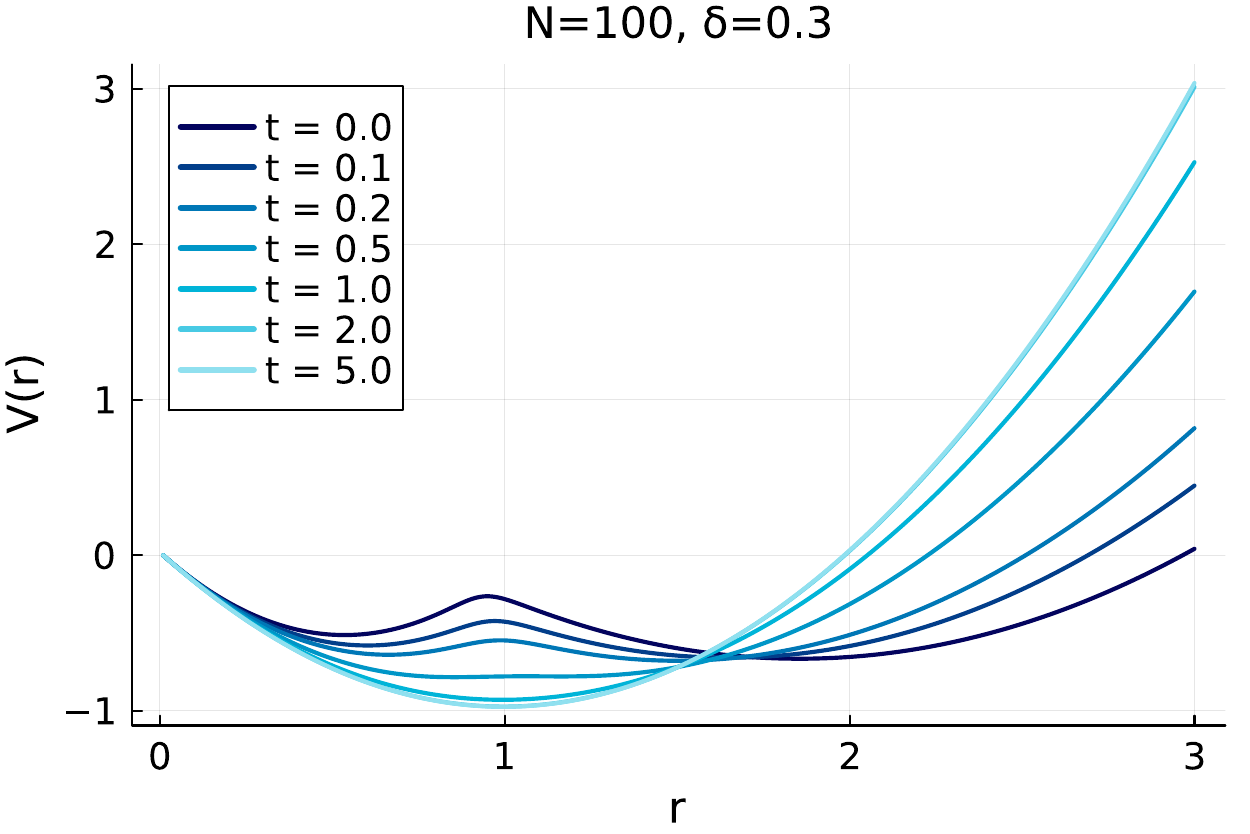"}
    \caption{Potential of the reverse SDE as a function of $r$ for different times for a mixture of two Gaussians with zero means and different isotropic variances $\sigma^2_{1,2} = 1 \pm \delta$. It is clearly noticeable a change in shape, from a single well for large $t$ to a double well for small $t$. We identify the speciation time in correspondence with the change in curvature in $r=1$.}
    \label{fig:potential}
\end{figure}

In this section, we study the case of a 2-Gaussians mixture in $\mathbb{R}^N$, both centered in $0$, with different isotropic variances $\sigma_1^2$ and $\sigma_2^2$. 
This setting reproduces the case of $a_{rs}=0$, for which the scaling of speciation time predicted by our criterion is $t\sim 1/4 \log N$, reported in Eq.~\ref{eq:asympt_spec_c}.
The distribution of the mixture at time $t$ is 
\begin{equation}
\label{eq:Gaussian-mixture}
    P_t(x) = \frac{1}{2} \frac{1}{(2\pi \Gamma_1)^{N/2}} e^{-\frac{\|x\|^2}{2\Gamma_1}} + \frac{1}{2} \frac{1}{(2\pi \Gamma_2)^{N/2}} e^{-\frac{\|x\|^2}{2\Gamma_2}},
\end{equation}
where $\Gamma_{1,2}(t)=\sigma_{1,2}^{2}e^{-2t}+(1-e^{-2t})$. For large $N$, the target probability density concentrates on thin shells of different radii depending on the two variances. We then choose $\sigma_{1}^{2}=1-\delta$ and $\sigma_{2}^{2}=1+\delta$, so $\Gamma_1(t) = 1 - \delta e^{-2t}$ and $\Gamma_2(t) = 1 + \delta e^{-2t}$, and tune the parameter $\delta>0$ in order to have two well distinct radii. 
The exact score function for this model is $S_t(x) = -\lambda(\|x\|, t) x$, where
\begin{equation}
    \lambda(\|x\|, t) = \frac{\Gamma_1^{-N/2} e^{-\|x\|^2/(2\Gamma_1)} / \Gamma_1 + \Gamma_2^{-N/2} e^{-\|x\|^2/(2\Gamma_2)} / \Gamma_2}{\Gamma_1^{-N/2} e^{-\|x\|^2/(2\Gamma_1)} + \Gamma_2^{-N/2} e^{-\|x\|^2/(2\Gamma_2)}},
\end{equation}
see \ref{app:Gaussian-mixture} for details. The reverse-time backward diffusion process for $x_t \in \mathbb{R}^N$ is again the Ornstein-Uhlenbeck process.
Introduce the scalar variable $r_t = \frac{\|x_t\|^2}{N}$, the reverse SDE for the radial coordinate reads
\begin{equation}
    dr = \left[2(r+1) - 4r\lambda(\sqrt{N r},t)\right] dt + 2 \sqrt{\frac{2r}{N}} dB_t = -\frac{\partial V_t(r)}{\partial r} dt + 2 \sqrt{\frac{2r}{N}} dB_t,
    \label{eq:radial-sde}
\end{equation}
where we have defined the effective potential $V_t(r)$ associated with the deterministic part of the SDE as minus the integral of the drift.
\begin{equation}
    V_t(r) = -\int_{0}^r \left[2(s + 1) - 4s\lambda(\sqrt{N s}, t)\right] ds
\end{equation}
In Fig.~\ref{fig:potential}, we can see how the shape of the potential evolves in time. One sees a symmetry-breaking phenomenon, from which one can estimate the speciation time for this model.
Indeed, we identify the transition as the time where the curvature vanishes in $r=1$: $\frac{\partial^2 V_t(r=1)}{\partial r^2} = 0$. Thus, we can derive a scaling for the speciation time by imposing that the derivative of $V_t(r)$ is zero at $r=1$, which translates into $4 - 4 \lambda(\sqrt{N}, t_s) = 0$, or more simply 
\begin{equation}
    \lambda(\sqrt{N}, t_s) = 1.
    \label{eq:speciation-gm}
\end{equation}
In \ref{app:verify-scaling} we verify that this condition is indeed satisfied for the desired scaling $t_s \sim \frac{1}{4}\log N$.

\section{Speciation times for multi-classes target distributions: 1D Ising mixtures}\label{sec:ising}
The aim of this section is to go beyond the toy model presented in the previous section, and focus on a structured high-dimensional probability distribution which leads to well-defined classes that cannot be identified from the first moments. 

We consider as target distribution a mixture of 1D Ising models at different inverse temperatures
\begin{equation}
    P(\sigma) = \sum_{r} w_r \frac{1}{Z(\beta_r)}e^{\beta_r\sum_i \sigma_i \sigma_{i+1}}.
\end{equation}
For large number of spins the different components of the probability distributions are peaked on different regions on the configuration space. However, first moments are identically zero at any $\beta$ for any component.

This example also showcases how our speciation criterion can be applied to the prediction of all speciation times, in the case of an arbitrary number of components. When the number of components is $n>2$, our method pinpoints a number of speciation transitions, of which only the first would have been otherwise computable. Notice that, similarly to the described in Sec.~\ref{sec:mixture-variance}, the components are not spatially well separated; the order parameter in this case is linked to the correlation length. Unlike the Gaussian mixture case, we cannot derive speciation explicitly from the potential. Thus, the general criterion of Eq.~\eqref{eq:speciation_crit} becomes essential to predict when components merge.
The Bayes attribution to component $r$ of a forward diffused sample $x_t$ reads in this case
\begin{equation}
P(s \mid x_t)=\frac{P(s)\sum_\sigma P(x_t \mid \sigma)P(\sigma \mid s)}{P(x_t)}=\frac{  \frac{w_s}{Z(\beta_s)}\sum_\sigma \ e^{\sum_i \frac{e^{-t}}{\Delta t} x_i^t \sigma_i +\beta_r\sum_i \sigma_i \sigma_{i+1}}}{\sum_c \frac{w_c}{Z(\beta_c)}\sum_\sigma \ e^{\sum_i \frac{e^{-t}}{\Delta t} x_i^t \sigma_i +\beta_c\sum_i \sigma_i \sigma_{i+1}} }
\end{equation}
The numerator is the partition function of a 1D Ising spin system at temperature $\beta_r$, subject to a (random) external field $x_t$. According to our criterion, speciations are marked by average difference between pairs of free energies
\begin{equation}
\label{eq:f_ising}
f_s(x, t)=\frac{1}{N}\log\left(\frac{1}{Z(\beta_s)}\sum_{\sigma}e^{\beta_{s}\sum_{i=1}^{N-1}\sigma_{i}\sigma_{i+1}+\frac{e^{-t}}{\Delta_{t}}\sum_{i=1}^{N-1}x_{i}\sigma_{i}}\right)
\end{equation}
becoming comparable with their fluctuations, when $x_t$ is generated by sampling a data point $a$ according to one of the $P_r$, and diffusing it forward in time. 
Even for relatively low values of $N$, one can obtain accurate speciation times by using the exact analytical expression for the average free energies and approximation \cref{eq:asympt_var} for the variance. The exact expression for the average free energies is derived in \ref{app:ising-free-ent} making use of replica computations, following the approach reported in~\cite{weigt1996replica}. The main difference is that in our case the individual features of the disorder $(x_t)_i$ are correlated, not i.i.d. variables. The values of $C_{rs}$ can be computed as shown in \ref{app:ising-crs}
\begin{equation}
C_{rs}
= 2\left[
     \frac{\tanh^2(\beta_r)}{1 - \tanh^2(\beta_r)}
   + \frac{\tanh^2(\beta_s)}{1 - \tanh^2(\beta_s)}
   - \frac{2\,\tanh(\beta_r)\,\tanh(\beta_s)}{1 - \tanh(\beta_r)\tanh(\beta_s)}
   \right].
\end{equation}
The speciation criterion reads
\begin{equation}
\label{eq:speciation_crit_approx}
    |f_{rr}(t_{rs})-f_{rs}(t_{rs})| = K \sqrt{\frac{C_{rs}}{2N}}e^{-2t_{rs}}.
\end{equation} 

Predicted speciation times will be compared to dynamical ones, defined experimentally in terms of U-turn experiments as follows:
sample an initial data $a$ from one of the $P_r$, and diffuse it forward to $x$, at time $t$. Then reverse the process, doing a ``U-turn'': start the backward process at time $t$ from $x$, and follow score-guided diffusion \cref{eq:score_guided_diffusion} for a time  $t'=t$, reaching $b$. The score (which can be computed using the transfer matrix method, see \ref{app:ising-exact-score}), ensures that $b$ is distributed according to Eq.~\eqref{eq:mixture}. We then assign $b$ to component $s$, by means of Bayes attribution. Repeating this experiment many times, one can monitor the probability that $b$ is found in component $s$ knowing that the initial condition $a$ was in component $r$. Misattribution probability remain very small until $t<t_{rs}$, and it grows when U-turn time surpasses speciation time.

\begin{figure}[t]
    \centering
    \includegraphics[width=0.5\linewidth]{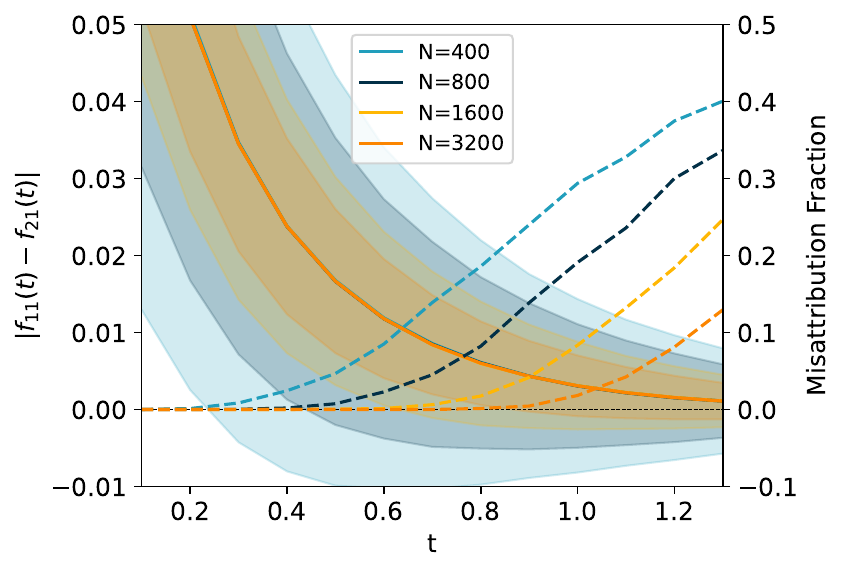}
    \caption{Analysis of a mixture of two 1D Ising models, at inverse temperatures $\beta_1=0.5$, $\beta_1=1$. Solid lines show the average free entropy difference for different number of spins $N$ as a function of forward diffusion time. The shading represents the $3\sigma$ confidence interval. Dashed lines show the misattribution fraction during the forward process. Misattribution starts to rise when zero enters the confidence interval. }
    \label{fig:free-entropy-forward}
\end{figure}

\subsection{2-Ising Mixture}
We begin by considering a mixture of two 1D Ising models with $\beta_1=0.5$ and $\beta_2=1.0$ and $w_1 = w_2 = 0.5$. In Fig.~\ref{fig:free-entropy-forward} we plot $f_{11} - f_{12}$ during the forward process (solid line) and three times free entropy difference variance as shading, for different number of spins. Dashed lines show empirical misattribution fractions for various $N$ values as a function of U-turn time, in the same color-coding. Empirical misattribution fractions are measured generating $10^4$ independent samples from each component, performing a U-turn and measuring the probability that the reconstructed data point is attributed to the same class as its origin. 
One can notice that when fluctuations of the free entropies difference cross zero, the misattribution fraction starts to rise.
Thus, we have an empirical confirmation that our criterion of Eq.~\eqref{eq:speciation_crit} captures the correct phenomenon: when the average free entropy difference is of the same order of its fluctuations, setting $K=3$, we find a misattribution probability of around $1\%$.  

\begin{figure}[b]
    \includegraphics[width=\linewidth]
    {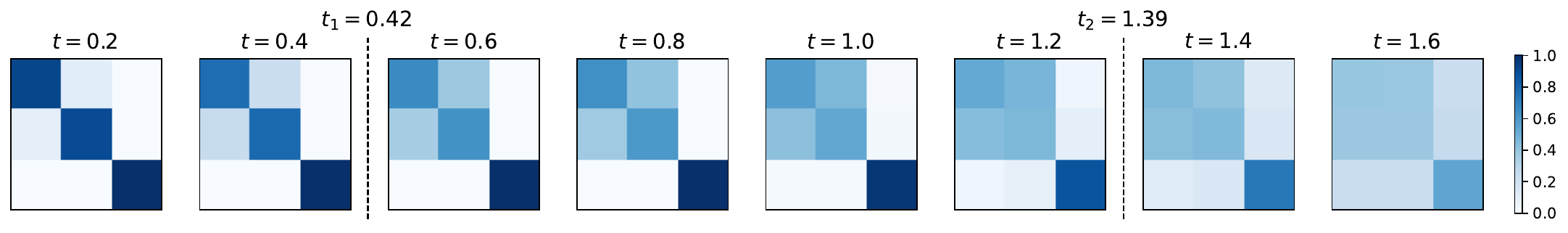}
    \caption{Attribution matrices at the end of the backward process for increasing U-turn times computed numerically with the transfer matrix method. Merging times predicted by our criterion are $t_1 \simeq 0.42$ for inverse temperatures $\beta_1$ and $\beta_2$, and $t_2 = 1.39$, for $\beta_2$ and $\beta_3$. The target distribution is a 3-Ising mixture with $\beta_1=0.2$, $\beta_2=0.3$, $\beta_3=1.0$; $N=1600$.}
    \label{fig:3-betas}
\end{figure}

\subsection{n-Ising Mixture} 
In this section, we consider mixtures of more than two Ising chains, showcasing how our theory can predict multiple speciation events. For an easy visual comparison between analytical speciation times and experimental attribution matrices (defined below) we set the constant $K=1$ in \cref{eq:speciation_crit_approx}, corresponding to a misattribution fraction $\sim 15\%$.

We start by considering a mixture of 3 Ising models with $\beta_1=0.2$, $\beta_2=0.3$, and $\beta_3=1.0$ and $w_1 = w_2 = w_3 = 1/3$. Sampling starting configurations from the mixture, we perform U-turn experiments at different times, and compute the probability of attributing the spin configuration reconstructed at the end of the backward process to any component.   
The resulting matrices are displayed in Fig.~\ref{fig:3-betas}, where the vertical dimension represents the origin component and the horizontal dimension represents the attribution component. Merging times predicted by our criterion are $t_1 \simeq 0.42$ for inverse temperatures $\beta_1$ and $\beta_2$, and $t_2 = 1.39$, for $\beta_2$ and $\beta_3$ (and, consequently, also $\beta_1$ and $\beta_3$ since the first two components were already merged). Notice how two values of $\beta$ are quite similar, leading to a short merging time. The empirical attribution matrices are almost diagonal for small $t_{U}$. As time increases, and in particular after passing the predicted merging times, the block structure in the attribution matrices is revealed.

\begin{figure}[t]
  \centering
      \begin{minipage}[t]{0.1\textwidth}
        \vspace{0pt}
        \includegraphics[height=0.22\textheight]{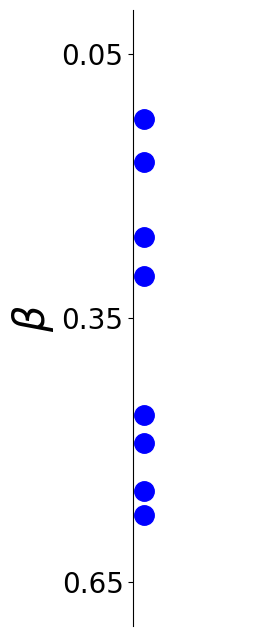}
      \end{minipage}
      \hfill
      \begin{minipage}[t]{0.88\textwidth} 
        \vspace{0pt}
        \includegraphics[width=\textwidth]{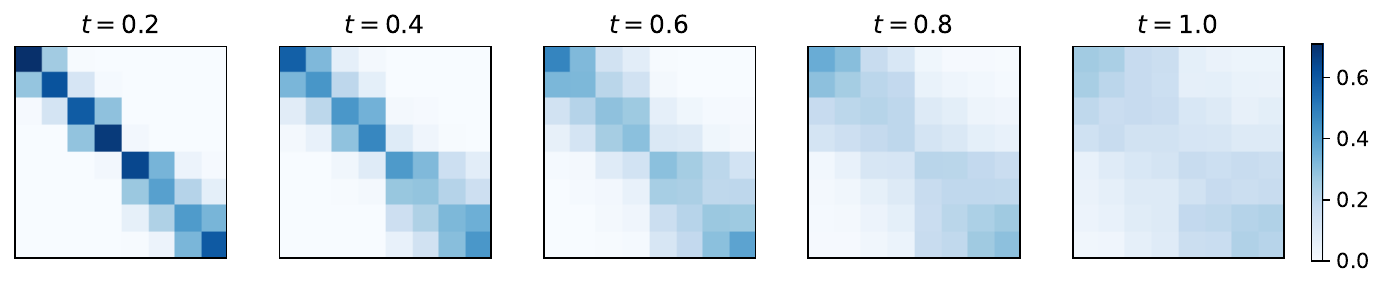}
        \includegraphics[width=0.94\textwidth]{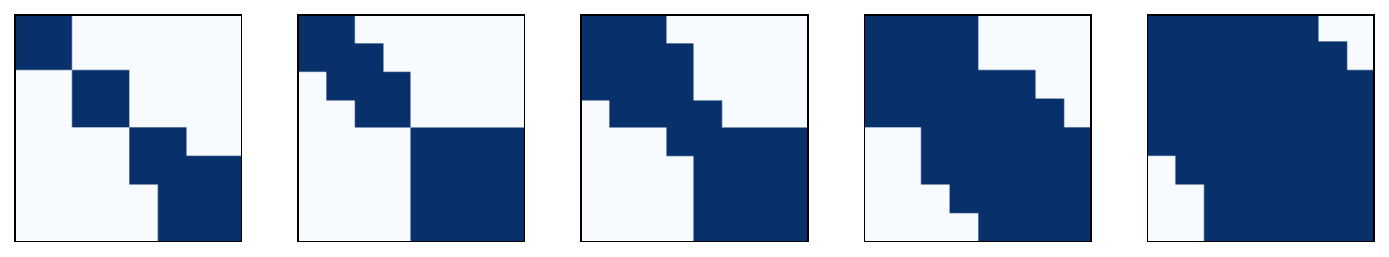}
      \end{minipage}%
      \caption{Attribution matrices at the end of the backward process for increasing U-turn times, computed numerically with the transfer matrix method (top row)  compared with the analytical ones (bottom row). The target distribution is a 8-Ising mixture with hierarchically generated temperatures (beta values displayed on the left). System size is $N=1600$.}
    \label{fig:hierarchical-betas}
\end{figure}

Next, we consider a homogeneous mixture of 8 Ising chains with hierarchically organized inverse temperatures ($\beta$ values displayed in \cref{fig:hierarchical-betas}). We perform again the U-turn experiment and obtain empirical the attribution matrices (top row). Empirical results are then compared with a matrix representation of the analytical predictions (bottom row), where blocks are colored if $|f_{rr}(t_{rs})-f_{rs}(t_{rs})| < \sqrt{\frac{C_{rs}}{2N}}e^{-2t_{rs}}$. There is good agreement between attribution matrices obtained analytically and numerically for all the U-turn times considered. 
The emerging block structure can again be explained with subsequent speciation events, where backwards trajectories commit to a smaller and smaller set of temperatures, and finally to a specific temperature. 

\section{Conclusions}
In this work, we have extended the theoretical understanding of speciation transitions in generative diffusion models to the case of data originating from mixture distributions, where components are peaked on different regions which cannot simply be identified by first moments. We have done so by introducing a new criterion for speciation, based on the analysis of the Bayes component attribution during the forward process, and free-entropy differences. The important outcome of our analysis is that merging times between different components scale logarithmically with the dimension of the sample space, and are separated by finite time gaps, giving rise to a hierarchical structure of attribution.

As a practical example, we applied the technique to mixtures of 1D Ising models, deriving explicit expressions for the mean and variance of the free entropy difference between components as a function of time. Our analysis reveals that both the mean and the standard deviation of this difference decay exponentially with time, and that the critical timescale for speciation is logarithmic in the system size, $t_S \sim \frac{1}{4} \log N$. 
We have validated our theoretical predictions with numerical experiments, demonstrating excellent agreement between analytical and empirical results for the onset of misattribution between components.
Our approach provides a principled criterion for speciation, that is applicable to a wide range of mixture statistical physics models, as long as each component has finite correlation length.

These findings contribute to the growing theoretical foundation of diffusion models and highlight the universality of the speciation phenomenon in high-dimensional generative processes. 
\\\\
{\bf Acknowledgments} 
GB was supported by ANR PRAIRIE-PSAI (France 2023) « ANR-23-IACL-0008».

\newcommand{\newblock}{}
\bibliographystyle{plainnat}
\bibliography{bibliography}

\newpage
\appendix
\section{Large time analysis}\label{app:large-time}
Explicitly, the average free entropy difference reads 
\begin{equation}
	f_{rr}(t)-f_{rs}(t) = \frac{1}{N}\left[\int dx P_{r}(x;t) \log  P_{r}(x;t)-\int dx P_{r}(x;t) \log  P_{s}(x;t)\right].
\end{equation}
Notice that this can be seen as the Kullback-Leibler divergence between the components of the mixture.
At large forward times one can obtain $P_{r}(x;t)$ by expanding in $e^{-2t}$ its exact expression. The result is a Gaussian distribution (see e.g. \cite{biroli2024dynamical})
\begin{equation}
\label{eq:logP}
\log P_r(x;t) = \text{const} + \frac{e^{-t}}{\Delta_t}\sum_{i=1}^N x_i \langle a_i \rangle_{r}
- \frac{1}{2\Delta_t}\sum_{i,j=1}^N x_i\, M_{r,ij}\, x_j + O\left((x e^{-t})^3\right),
\end{equation}
where
\begin{equation}
M_{r,ij} = \delta_{ij} - e^{-2t}\left( \langle a_i a_j \rangle_{r} - \langle a_i \rangle_{r} \langle a_j \rangle_{r} \right).
\end{equation}
and $\Delta_t = 1-e^{-2t}$, so when we expand in $e^{-t}$ we write
\begin{equation}
	\frac{1}{\Delta_t}=\frac{1}{1-e^{-2t}}=1+e^{-2t}+e^{-4t}+O(e^{-4t})
	\qquad
	\frac{1}{\Delta_t^2}=\frac{1}{(1-e^{-2t})^2}=1+2e^{-2t}+3e^{-4t}+O(e^{-4t})
\end{equation}
Completing the square in \eqref{eq:logP} yields a Gaussian with
\begin{align}
	\mu_{r}(t) &= \frac{e^{-t}}{\Delta_t}\,M_{r}^{-1}\langle a\rangle_{r}
	= e^{-t}\left(1+e^{-2t}+e^{-4t}\right)\left[I+e^{-2t} C+O(e^{-4t})\right]\langle a\rangle_{r} \\
	&= e^{-t}\left[\langle a\rangle_{r} + e^{-2t}\left(\langle a\rangle_{r} + C_{r}\langle a\rangle_{r}\right) + O(e^{-4t})\right],\\
	\Sigma_{r}(t) &= \frac{1}{\Delta_t} \Delta_t M_{r}^{-1} = I+e^{-2t} C_{r}+O(e^{-4t}),
\end{align}
where $C_{r,ij} = \langle a_i a_j \rangle_{r} - \langle a_i \rangle_{r} \langle a_j \rangle_{r} $.
Then, the average free entropy difference is a Kullback-Leibler divergence between Gaussians
\begin{equation}
	D_{\mathrm{KL}}(\mathcal N(\mu_r,\Sigma_r)\,\|\,\mathcal N(\mu_s,\Sigma_s))
	=\frac12 \left[ \mathrm{Tr}(\Sigma_s^{-1}\Sigma_r) + (\mu_s-\mu_r)^\top \Sigma_s^{-1} (\mu_s-\mu_r) - N - \log\frac{\det \Sigma_r}{\det \Sigma_s} \right].
\end{equation}
Define
\begin{align}
\Delta_{rs} a_i & = \langle a_i \rangle_r - \langle a_i \rangle_s, \\
\Delta_{rs} C_{ij} &= \left( \langle a_i a_j \rangle - \langle a_i \rangle \langle a_j \rangle \right)_r
- \left( \langle a_i a_j \rangle - \langle a_i \rangle \langle a_j \rangle \right)_s, 
\end{align}
then using the Gaussian KL formula and expanding up to $O(e^{-4t})$ gives
\begin{align}
&\frac{1}{N} D_{\mathrm{KL}}\left(P_r(\cdot,t)\,\|\,P_s(\cdot,t)\right)
= \frac{1}{2N}\frac{e^{-2t}}{\Delta_t}\sum_i (\Delta_{rs} a_i)^2 + \frac{e^{-4t}}{4N}\frac{1}{\Delta_t^{2}}\sum_{i,j}\left(\Delta_{rs} C_{ij}\right)^2 \\
&+ \frac{e^{-4t}}{2N}\frac{1}{\Delta_t}\left[
2\sum_i \Delta_{rs} a_i \left(\sum_j C_{s,ij}\langle a_j\rangle_s - \sum_j C_{r,ij}\langle a_j\rangle_r \right)
- \sum_{i,j} \Delta_{rs} a_i C_{s,ij} \Delta_{rs} a_j
\right] + O(e^{-4t}).
\label{eq:KL}
\end{align}
Now expand $\Delta_t$ and retain terms through $O(e^{-4t})$

\begin{align}
\frac{1}{N} D_{\mathrm{KL}} &= 
\frac{1}{2N}\left(e^{-2t}+e^{-4t}\right)\sum_i (\Delta_{rs} a_i)^2
+ \frac{e^{-4t}}{4N}\sum_{i,j}\left(\Delta_{rs} C_{ij}\right)^2\\
&+ \frac{e^{-4t}}{2N}\left[
2\sum_i \Delta_{rs} a_i \left(\sum_j C^{(s)}_{ij}\,\langle a_j\rangle_s - \sum_j C^{(r)}_{ij}\langle a_j\rangle_r \right)
- \sum_{i,j} \Delta_{rs} a_i C^{(s)}_{ij}\Delta_{rs} a_j
\right]+O(e^{-4t}) \\
&\simeq \frac{a_{rs}}{2} (e^{-2t} + e^{-4t}) + \frac{C_{rs}}{4} e^{-4t} + e^{-4t} S_{rs}, 
\end{align}
where 
\begin{align}
    a_{rs} &= \frac{\sum_i (\Delta_{rs} a_i)^2}{N}, \qquad C_{rs} =\frac{ \sum_{i,j}\left(\Delta_{rs} C_{ij}\right)^2 }{N}, \\
    S_{rs} &=\frac{2\sum_i \Delta_{rs} a_i \left(\sum_j C_{s,ij}\,\langle a_j\rangle_s - \sum_j C_{r,ij}\langle a_j\rangle_r \right)- \sum_{i,j} \Delta_{rs} a_i C_{s,ij}\Delta_{rs} a_j}{2N}
\end{align}

Leveraging the Gaussian approximation for $P_r(x;t)$ at large $t$, one can also approximate $\mathrm{Var}[\frac{1}{N}\log  P_s(x;t)-\frac{1}{N} \log  P_r(x;t)]$. Specifically:
\begin{equation}
\frac{1}{N}\log P_r(x;t) - \frac{1}{N}\log P_s(x;t)
= \frac{1}{N}\left[
\frac{e^{-t}}{\Delta_t}\sum_i x_i\,\Delta_{rs} a_i
+ \frac{e^{-2t}}{2\Delta_t}\sum_{i,j} x_i\,\Delta_{rs} C_{ij}\,x_j
\right] + O\left((x e^{-t})^3\right).
\end{equation}
The dominant contribution to the variance comes from considering $x_i\simeq z_i$ with $z_i\stackrel{\mathrm{i.i.d.}}{\sim}\mathcal N(0,1)$. 
The linear and quadratic parts are uncorrelated for a centered Gaussian, so
\begin{align}
\mathrm{Var}\left[\frac{1}{N}\log  P_s(x;t)-\frac{1}{N} \log  P_r(x;t)\right] 
&= \mathrm{Var}\left[\frac{e^{-t}}{N\Delta_t}\sum_i \Delta_{rs} a_i z_i\right]
+ \mathrm{Var}\left[\frac{e^{-2t}}{2N\Delta_t}\sum_{i,j}\Delta_{rs} C_{ij} z_i z_j\right] + O(e^{-6t}) \\
&= \frac{e^{-2t}}{N^2}\frac{1}{\Delta_t^{2}}\sum_i (\Delta_{rs} a_i)^2
+ \frac{e^{-4t}}{2N^2}\frac{1}{\Delta_t^{2}}\sum_{i,j} (\Delta_{rs} C_{ij})^2 + O(e^{-6t}).
\end{align}
Expanding also $\Delta_t$ and keeping through $O(e^{-4t})$,
\begin{align}
\label{eq:large_time_Var}
\mathrm{Var}\left[\frac{1}{N}\log P_r(\mathbf{x},t) - \frac{1}{N}\log P_s(\mathbf{x},t)\right]
&\simeq\left(e^{-2t} + 2e^{-4t}\right)\frac{1}{N^2}\sum_i (\Delta_{rs} a_i)^2
+\frac{e^{-4t}}{2N^2}\sum_{i,j} \left(\Delta_{rs} C_{ij}\right)^2 \\
&\simeq \frac{a_{rs}}{N} (e^{-2t} + 2e^{-4t}) + \frac{C_{rs}}{2N}e^{-4t}.
\end{align}

In this large $t$ regime, the criterion for speciation then reads
\begin{enumerate}
    \item If $a_{rs} \ne 0$, 
	\begin{equation}
		\frac{a_{rs}}{2} e^{-2t} = K\cdot \sqrt{\frac{a_{rs}}{N}} e^{-t} \implies t_{rs}= \frac12 \log N -\frac{1}{2}\log \left(\frac{4}{a_{rs}}\right) -  \log K.
	\end{equation}
    This recovers the speciation time scaling obtained in \cite{biroli2024dynamical}.
    \item If $a_{rs} = 0$,
	\begin{equation}
		\frac{C_{rs}}{4}e^{-4t} = K\cdot \sqrt{\frac{C_{rs}}{2 N}} e^{-2t}  \implies t_{rs}=\frac{1}{4}\log N -\frac{1}{4}\log\left(\frac{8}{C_{rs}}\right) - \frac{\log K}{2}.
	\end{equation}
    This case extends previous literature on speciation time to cases where the class distribution do not have any first moment.
\end{enumerate}
Notice that, on this timescale, $\mathrm{Var}\left[ 1/N \log  P_r(x;t)- 1/N \log  P_s(x;t) \right]=O(\frac{1}{N^2})$: this is due to a combination of factors: a free entropy variance has a natural scaling of $1/N$, at fixed time. But speciation happens on a timescale which is logarithmic in $N$, which contributes an additional $1/N$ factor to the variance. Hence, an equivalent definition of speciation time is $f_{rr}(t)-f_{rs}(t) \simeq 1/N$.

\section{Mixture of Gaussians with different covariance}\label{app:Gaussian-mixture}
\subsection{General criterion}
For completion, we also compute the speciation time prediction from \cref{eq:speciation_crit} for the mixture of Gaussians defined in Eq.\eqref{eq:Gaussian-mixture}. The left-hand side is
\begin{align}
    f_{11}(t) - f_{21}(t) = -\frac{1}{2} \left[ \log \frac{\Gamma_1(t)}{\Gamma_2(t)} +1 - \frac{\Gamma_1(t)}{\Gamma_2(t)} \right]
\end{align}
and the right-hand side 
\begin{align}
    \mathrm{Var}\left[ \frac{1}{N}\log P_1(x;t) - \frac{1}{N}\log P_2(x;t)\right] = \frac{1}{2N} \Gamma_1(t)^2 \left( \frac{1}{\Gamma_1(t)} - \frac{1}{\Gamma_2(t)}\right)^2.
\end{align}
The Bayes attribution criterion then reads
\begin{equation}
    \frac{1}{2} \left[ \log \frac{\Gamma_1(t)}{\Gamma_2(t)} +1 - \frac{\Gamma_1(t)}{\Gamma_2(t)} \right] = \frac{1}{\sqrt{2N}} \left(1 - \frac{\Gamma_1(t)}{\Gamma_2(t)}\right). 
\end{equation}
For our choice of $\Gamma(t)$
\begin{equation}
    \frac{\Gamma_1(t)}{\Gamma_2(t)} = \frac{1-\delta e^{-2t}}{1+\delta e^{-2t}} \approx 1 - 2\delta e^{-2t}
\end{equation}
so 
\begin{align}
    \log(1-2\delta e^{-2t}) + 2\delta e^{-2t} \simeq \frac{2\delta e^{-2t}}{\sqrt{2N}}
\end{align}
and expanding the log we obtain
\begin{equation}
    \delta e^{-2t} \simeq \frac{1}{\sqrt{2N}}
\end{equation}
from which $t_s = \frac{1}{4}\log N$.

\subsection{Score function}
To compute the exact score function, defined as $S_t(x) = \nabla_x \log p_t(x)$, one can define the weights:
\begin{equation}
    w_i(x) = \Gamma_i^{-N/2} e^{-\frac{\|x\|^2}{2\Gamma_i}}
\end{equation}
and call
\begin{equation}
    m_i(x) = \frac{w_i(x)}{w_1(x) + w_2(x)}.
\end{equation}

Then, the exact score is:
\begin{align}
    S_t(x) &= -x \left( \frac{m_1(x)}{\Gamma_1} + \frac{m_2(x)}{\Gamma_2} \right) \\
    &= -\lambda(\|x\|, t) x
\end{align}
where
\begin{equation}
    \lambda(\|x\|, t) = \frac{\Gamma_1^{-N/2} e^{-\|x\|^2/(2\Gamma_1)} / \Gamma_1 + \Gamma_2^{-N/2} e^{-\|x\|^2/(2\Gamma_2)} / \Gamma_2}{\Gamma_1^{-N/2} e^{-\|x\|^2/(2\Gamma_1)} + \Gamma_2^{-N/2} e^{-\|x\|^2/(2\Gamma_2)}}
\end{equation}

\subsection{Radial SDE}
We consider the Ornstein-Uhlenbeck reverse-time diffusion process for $x_t \in \mathbb{R}^N$:
\begin{equation}
    dx = \left(x + 2 S_t(x) \right) dt + \sqrt{2} dW_t
\end{equation}
where $dW_t$ is standard Brownian motion. 
Our goal is to obtain an equation for the speciation time, similarly to what was done in \cite{biroli2024dynamical} for the case of a mixture of Gaussians centered respectively in $\pm m$, with $\|m\|^2 = N\mu $.
Define the scalar variable:
\begin{equation}
    r_t = \frac{\|x_t\|^2}{N}
\end{equation}

Using It\^{o}'s lemma:
\begin{equation}
    dr = \frac{2}{N} x^\top dx + \frac{1}{N} \mathrm{Tr}(dx \ dx^\top)
\end{equation}
From the SDE for $x_t$ we see $dx \ dx^\top = 2 I dt$, so
\begin{align}
    dr &= \frac{2}{d} x^\top (x + 2 S_t(x)) dt + 2 dt + \frac{2\sqrt{2}}{N} x^\top dW_t \\
    &= \left[2r + 4 \alpha(t,x) + 2\right] dt + \frac{2\sqrt{2}}{N} x^\top dW_t
\end{align}
where $\alpha(t,x) = \frac{1}{N} x^\top S_t(x)$.

Using our expression for the score $S_t(x) = -\lambda(\|x\|,t) x$, we get:
\begin{equation}
    \alpha(r,t) = -r \lambda(\sqrt{N r}, t)
\end{equation}
so the drift becomes:
\begin{equation}
    2r + 4\alpha(r,t) + 2 = 2(r + 1) - 4 r \lambda(\sqrt{N r}, t)
\end{equation}

The noise term can be rewritten as 
\begin{equation}
    \frac{2\sqrt{2}}{N} x^\top dW_t \approx 2 \sqrt{\frac{2r}{N}} dB_t
\end{equation}
which is negligible for large $N$. 

Putting all together, we find
\begin{equation}
    dr = \left[2(r+1) - 4r\lambda(\sqrt{N r},t)\right] dt + 2 \sqrt{\frac{2r}{N}} dB_t
    \label{eq:radial-sde-app}
\end{equation}

\subsection{Scaling of speciation time}\label{app:verify-scaling}

We can verify that the scaling for the speciation time obtained from the change if shape of the potential is the expected one of $t_s \sim 1/4 \log N$.
To do so, consider
	\begin{equation}
		\lambda(\sqrt{N},t)
		=
		\frac{
			\Gamma_1^{-N/2-1} \exp\!\big(-\tfrac{N}{2\Gamma_1}\big)
			+\Gamma_2^{-N/2-1} \exp\!\big(-\tfrac{N}{2\Gamma_2}\big)
		}{
			\Gamma_1^{-N/2} \exp\!\big(-\tfrac{N}{2\Gamma_1}\big)
			+\Gamma_2^{-N/2} \exp\!\big(-\tfrac{N}{2\Gamma_2}\big)
		},
	\end{equation}
	with
	\begin{equation}
		\Gamma_1 = 1 - \varepsilon,\qquad \Gamma_2 = 1 + \varepsilon,
		\qquad \varepsilon = \delta e^{-2t}.
	\end{equation}
	
	We are interested in the scaling $t_s = \frac{1}{4}\log N + b$, so that
	\begin{equation}
		\varepsilon_s = \delta e^{-2t_s}
		= \frac{\delta e^{-2b}}{\sqrt{N}}
		= \frac{a}{\sqrt{N}},
		\qquad a := \delta e^{-2b},
	\end{equation}
	and we write simply $\varepsilon = a/\sqrt{N}$.
	
	Define
	\begin{equation}
		B_1 = \Gamma_1^{-N/2} \exp\!\Big(-\frac{N}{2\Gamma_1}\Big),\qquad
		B_2 = \Gamma_2^{-N/2} \exp\!\Big(-\frac{N}{2\Gamma_2}\Big),
	\end{equation}
	so that $\lambda=\frac{B_1/\Gamma_1 + B_2/\Gamma_2}{B_1 + B_2}$.
	One can factor out $B_1$ and set $r = \frac{B_2}{B_1}$, to get
	\begin{equation}
		\lambda
		=
		\frac{\Gamma_1^{-1} + r\,\Gamma_2^{-1}}{1 + r}
		=
		\frac{(1-\varepsilon)^{-1} + r\,(1+\varepsilon)^{-1}}{1 + r}.
	\end{equation}
	
A standard Taylor expansion in $\varepsilon$ of $r$ gives $\log r=\frac{2N}{3}\,\varepsilon^3 + O(N\varepsilon^5)$, and with $\varepsilon = a/\sqrt{N}$ this becomes
	\begin{equation}
		\Delta = \log r
		= \frac{2}{3}\frac{a^3}{\sqrt{N}} + O\Big(\frac{1}{N^{3/2}}\Big).
	\end{equation}
	Thus $r = e^{\Delta}= 1 + \Delta + \frac{\Delta^2}{2} + O(\Delta^3)$, and both $\Delta$ and $\varepsilon$ are of order $N^{-1/2}$.

    Using the expansions above and keeping terms up to second order in $\varepsilon$ and $\Delta$,

	\begin{equation}
		\lambda
		\simeq
		\frac{
			1 + \varepsilon^2 + \frac{\Delta}{2}(1-\varepsilon) + \frac{\Delta^2}{4}
		}{
			1 + \frac{\Delta}{2} + \frac{\Delta^2}{4}
		}.
	\end{equation}
	
    Now expand the ratio for small $\Delta$ up to terms of order $O(\varepsilon^2\Delta)$ and $O(\Delta^2)$,
	\begin{align}
		\lambda\simeq 1 + \varepsilon^2 - \frac{\Delta\varepsilon}{2}
		+ O\Big(\frac{1}{N^{3/2}}\Big).
	\end{align}
	
	Finally, we insert the scalings for $\varepsilon$ and $\Delta$
	\begin{equation}
		\lambda(\sqrt{N}, t_s)
		= 1 + \frac{a^2}{N} - \frac{a^4}{3N}
		+ O\Big(\frac{1}{N^{3/2}}\Big)
		= 1 + \frac{a^2\Big(1 - \frac{a^2}{3}\Big)}{N}
		+ O\Big(\frac{1}{N^{3/2}}\Big).
	\end{equation}
	
	To cancel the $1/N$ term we impose
	\begin{equation}
		a^2\Big(1 - \frac{a^2}{3}\Big) = 0,
	\end{equation}
	which gives the nontrivial solution $a^2 = 3$, i.e. $a = \sqrt{3}$.
	Since $a = \delta e^{-2b}$, we have
	\begin{equation}
		\delta e^{-2b} = \sqrt{3}
		\quad\Rightarrow\quad
		b = \frac12\log\delta - \frac14\log 3,
	\end{equation}
	and thus
	\begin{equation}
		t_s = \frac14\log N + \frac12\log\delta - \frac14\log 3.
	\end{equation}

\section{1D Ising mixtures}\label{app:ising-model}
This section details the computation of the score, the empirical free entropy and the average free entropy in the case where $P(a)=\sum_{r=1}^R w_r P_{\beta_r}(a)$ is a mixture of 1D Ising models at different inverse temperatures $\beta_{r}$, where every component of the mixture has the form
\begin{equation}
    P_{\beta}(s)=\frac{1}{Z(\beta)}e^{\beta\sum_{i=1}^{N-1}\sigma_{i}\sigma_{i+1}},
\end{equation}
where $Z(\beta)=(2\cosh\beta)^{N-1}$, and $\sum_r w_r=1$ . 

\subsection{Bayesian Attribution}\label{app:bayes-attr}
The Bayesian probability for a trajectory to have originated from component
$s$ given its value $x_{t}$ is given by
\begin{equation}
P(s \mid x_{t})=\frac{P(s,x_{t})}{P(x_{t})}=\frac{P(s)P(x_{t} \mid s)}{P(x_{t})}=\frac{P(s)\sum_\sigma P(x_{t} \mid \sigma)P(\sigma \mid s)}{P(x_{t})}
\end{equation}
leading to
\begin{align}
P(s\mid x_t) &= \frac{  \frac{w_s}{Z(\beta_s)}\sum_\sigma \ e^{\sum_i \frac{e^{-t}}{\Delta t} x_i^t \sigma_i +\beta_r\sum_i \sigma_i \sigma_{i+1}}}{\sum_c \frac{w_c}{Z(\beta_c)}\sum_\sigma \ e^{\sum_i \frac{e^{-t}}{\Delta t} x_i^t \sigma_i +\beta_c\sum_i \sigma_i \sigma_{i+1}} } \\
&=\frac{w_{s}e^{Nf_s(x, t)-\log Z(\beta_{s})}}{\sum_c  w_{c}e^{NP_c(x, t)(x;t)-\log Z(\beta_{c})}  }
\end{align}
where
\begin{equation}\label{eq:free-ent}
f_s(x, t)=\frac{1}{N}\log\left(\sum_{\sigma}e^{\beta_{s}\sum_{i=1}^{N-1}\sigma_{i}\sigma_{i+1}+\frac{e^{-t}}{\Delta_{t}}\sum_{i=1}^{N-1}x_{i}\sigma_{i}}\right)
\end{equation}
is the free-entropy of a 1D Ising model with random field $h_i = \frac{e^{-t}}{\Delta_t}x_i$. It can be computed using the transfer matrix method.

\subsection{Analytic computation of the average free entropy}\label{app:ising-free-ent}
Our criterion for speciation prescribes computing the average $f_{rs}$ \cref{eq:free-ent}, when $x$ is generated by sampling a data point $a$ according to  $P_r$, and diffusing it forward in time. Then, $x= a e^{-t}+z\sqrt{\Delta_t}$ where $z$ is a centered $N$-dimensional Gaussian random variable with covariance $\mathbb{I}_N$. We can obtain an exact expression for this quantity, making use of replica computations following the approach reported in~\cite{weigt1996replica}. The main difference is that in our case the noise is partially correlated with the data. Hence
\begin{equation}
    f_{rs}(t)= \left\langle \frac{1}{N}\log\left(\sum_{\sigma}e^{\beta_{r}\sum_{i=1}^{N-1}\sigma_{i}\sigma_{i+1}+\frac{e^{-t}}{\Delta_{t}}\sum_{i=1}^{N-1}\left(a_{i}e^{-t}+z_{i}\sqrt{\Delta_{t}}\right)\sigma_{i}}\right) \right\rangle_{a\sim P_r(a), z}
\end{equation}
We compute this quantity by means of the replica trick, in the Replica Symmetric approximation. First, we need to compute integer powers 
\begin{align}
    \left\langle Z^n \right\rangle  &=\left\langle \left(\sum_{\sigma}e^{\beta_{r}\sum_{i=1}^{N-1}\sigma_{i}\sigma_{i+1}+\frac{e^{-t}}{\Delta_{t}}\sum_{i=1}^{N-1}\left(a_{i}e^{-t}+z_{i}\sqrt{\Delta_{t}}\right)\sigma_{i}}\right)^n \right\rangle_{a\sim P_r(a), z} \\
    &= \frac{1}{Z(\beta_{r})}\sum_{{\sigma^{a}}}e^{\beta_{r}\sum_{i}\sigma_{i}^{0}\sigma_{i+i}^{0}+\beta_s\sum_{i,a}\sigma_{i}^{a}\sigma_{i+i}^{a}+\gamma^{2}\sum_{i}\sigma_{i}^{0}\sum_{a}\sigma_{i}^{a}}\langle\langle e^{\gamma\sum_{i}z_{i}\sum_{a}\sigma_{i}^{a}}\rangle\rangle_{z}.
\end{align}
where we defined $\gamma=\frac{e^{-t}}{\sqrt{\Delta_{t}}}$. The $n$ replicas $\sigma^{a}$, $a=1,\ldots,n$ account for the power $n$, the $n+1$ replica $\sigma^{0}$ accounts for the prior $P_r(a)=\frac{1}{Z(\beta_r)}e^{\beta_r\sum_{i}a_{i}a_{i+1}}$, and then make the continuation to real $n$ using the identity\begin{equation}
    f_{rs}(t)=\lim_{n\to 0} \frac{\langle Z^n \rangle -1}{n}.
\end{equation}
To describe the replica symmetric subspace we introduce vectors $|\sigma^{0},a_{1}\ldots a_{p}\rangle$, where we have $p$ spins
up at indices $a_{j}$. Then, we take the sum of such vectors
\begin{equation}
||\sigma^{0},p\rangle\rangle=\sum_{a_{1}<\ldots<a_{p}}|\sigma^{0},a_{1}\ldots a_{p}\rangle
\end{equation}
and the collection of these vectors ${||\sigma^{0},p\rangle\rangle}_{\sigma^{0}=\pm1,p=1,\ldots,n}$
gives the RS subspace of dimension $2(n+1)$ (see \cite{lucibello2014onedimensional} for details). Then, we look at the
replicated transfer matrix $T_{2^{n+1}\times2^{n+1}}$ projected onto
this subspace
\begin{equation}
\langle\langle\sigma^{0},q||T||\tilde{\sigma}^{0},p\rangle\rangle=e^{\beta_r\sigma^{0}\tilde{\sigma}^{0}}e^{\gamma^{2}\sigma^{0}(2q-n)}\langle\langle e^{\gamma z(2q-n)}\rangle\rangle_{z}\sum_{r=r_{min}}^{r_{max}}\binom{q}{r}\binom{n-q}{p-r}e^{\beta_s(4r-2q-2p+n)}
\end{equation}
with $r_{min}=\max(0,p+q-n)$ and $r_{max}=\min(p,q)$. The site-dependent partition function in the RS approximation is given by
\begin{equation}
\label{eq:iterative}
Z_{i+1}(\tilde{\sigma}^{0},p)=\sum_{\sigma^{0}=\pm1}\sum_{q=0}^{n}T(\sigma^{0},q;\sigma^{0},p)Z_{i}(\sigma^{0},q).
\end{equation}
We can write this in a function basis, defining $Z_{i}^{\sigma^{0}}[x] =\sum_{p=0}^{n} Z_{i}(\sigma^{0},p)x^{p}$.
\Cref{eq:iterative} now reads
\begin{equation}
    Z_{i+1}^{\tilde{\sigma}^{0}}[x] =\int_{0}^{\infty} \sum_{\sigma^{0}=\pm1}e^{\beta_r\sigma^{0}\tilde{\sigma}^{0}} K_{n}^{\sigma^{0}}(x,y)Z_{i}^{\sigma^{0}}[y] dy
\end{equation}
where the kernel
$K_{n}^{\sigma^{0}}(x,y)=e^{\beta_s n}(1+xe^{-2\beta_s})^{n}e^{-\gamma^{2}\sigma^{0}n}\langle\langle\delta\left(y-g^{\sigma^{0}}(x,z)\right)e^{-n\gamma z}\rangle\rangle_{z}$ is the representation of the transfer matrix in the space of polynomials, and 
\begin{equation}
g^{\sigma^{0}}(x,z)=e^{2\gamma z}\frac{e^{-\beta_s}+xe^{\beta_s}}{e^{\beta_s}+xe^{-\beta_s}}e^{2\gamma^{2}\sigma^{0}}.
\end{equation}
As we are interested in the behavior of the largest eigenvalue of the transfer matrix kernel in the limit $n\to 0$, we focus on the eigenvectors with largest eigenvalue of the $n=0$ kernel 
\begin{equation}
K=\left(\begin{array}{cc}
\frac{e^{\beta_r}}{2\cosh\beta_r}K^{+}(x,y) & \frac{e^{-\beta_r}}{2\cosh\beta_r}K^{-}(x,y)\\
\frac{e^{-\beta_r}}{2\cosh\beta_r}K^{+}(x,y) & \frac{e^{\beta_r}}{2\cosh\beta_r}K^{-}(x,y)
\end{array}\right),
\end{equation}
where 
\begin{equation}
K^{\sigma^{0}}(x,y)=\langle\langle\delta\left(y-g^{\sigma^{0}}(x,z)\right)\rangle\rangle_{z}.
\end{equation}
We denote with $\Psi(x)=\left(\begin{array}{c}
\psi^{+}(x)\\
\psi^{-}(x)
\end{array}\right)$ and $\Phi(x)=\left(\begin{array}{c}
\phi^{+}(x)\\
\phi^{-}(x)
\end{array}\right)$ the right and left eigenvectors respectively.
It is actually easy to see that at leading order in $n$ the largest  eigenvalue is $1$ and 
$\Psi(x)=\left(\begin{array}{c}
1\\
1
\end{array}\right)$, while $\Phi(x)$ can be computed numerically, by iteratively applying $K$ to any sufficiently well-behaved function, and normalized so that $\int \phi^{+,-}(x)dx=1$. To linear order in $n$, the largest eigenvalue $\lambda=1+kn$ is found by expanding the Kernel
\begin{align}
K_{n}^{\sigma^{0}}(x,y) & \approx\langle\langle\delta\left(y-g^{\sigma^{0}}(x,z)\right)\rangle\rangle_{z}+n\langle\langle\delta\left(y-g^{\sigma^{0}}(x,z)\right)\left[-\gamma z-\gamma^{2}\sigma^{0}+\beta_s+\log(1+bx)\right]\rangle\rangle_{z}\\
 & =K_{0}^{\sigma^{0}}(x,y)+n\delta K^{\sigma^{0}}(x,y)
\end{align}
and leveraging the eigenvalue condition
\begin{align*}
k & =\int dx\,dy\ \Phi(x)\delta K(x,y)\Psi(y)\\
 & =\int dx\ \phi^{+}(x)\left(\frac{e^{\beta_r}}{2\cosh\beta_r}\delta K^{+}(x)+\frac{e^{-\beta_r}}{2\cosh\beta_r}\delta K^{+}(x)\right)\\
 &+\int dx \phi^{-}(x)\left(\frac{e^{-\beta_r}}{2\cosh\beta_r}\delta K^{-}(x)+\frac{e^{\beta_r}}{2\cosh\beta_r}\delta K^{-}(x)\right)\\
 & =\frac{1}{2}\int dx\ [\phi^{+}(x)\delta K^{+}(x)+\phi^{-}(x)\delta K^{-}(x)]
\end{align*}
where $\delta K^{\sigma^{0}}(x)=\beta_s-\gamma^{2}\sigma^{0}+\log(1+xb)$.
Finally, defining $\hat{\phi}^{+,-}(t):=e^t\phi^{+,-}(e^t)$, we obtain
\begin{equation}
k=\frac{\left[\beta_s-\gamma^{2}+\int dt\ \log(1+be^{t})\hat{\phi}^{+}(t)\right]+\left[\beta_s+\gamma^{2}+\int dt\ \log(1+be^{t})\hat{\phi}^{-}(t)\right]}{2},
\end{equation}
and 
\begin{equation}
    f_{rs}(t)=\lim_{n\to 0} \frac{\langle Z^n \rangle -1}{n}=k.
\end{equation}

\subsection{Computation of $C_{rs}$}\label{app:ising-crs}
For inverse temperatures $\beta_r$ and $\beta_s$ we define $t_r = \tanh(\beta_r)$, $t_s = \tanh(\beta_s)$.
In zero field the magnetization vanishes and the two-point function in the
thermodynamic limit is
\begin{equation}
\langle \sigma_i \sigma_j \rangle_r = t_r^{\,|i-j|},
\end{equation}
thus the connected correlations are
\begin{equation}
C^{(r)}_{ij} = t_r^{\,|i-j|}, 
\qquad 
C^{(s)}_{ij} = t_s^{\,|i-j|}.
\end{equation}
We therefore have
\begin{equation}
\Delta_{rs} C_{ij}
= C^{(r)}_{ij} - C^{(s)}_{ij}
= t_r^{\,|i-j|} - t_s^{\,|i-j|}.
\end{equation}

The quantity of interest is
\begin{equation}
C_{rs}
= \frac{1}{N} \sum_{i,j=1}^N 
   \left( \Delta_{rs} C_{ij} \right)^2 .
\end{equation}

In a periodic chain, for any fixed separation $k\ge 1$ there are $2N$
ordered pairs $(i,j)$ with $|i-j|=k$, while the $k=0$ term vanishes because
$\Delta_{rs}C_{ii}=0$.  
Taking the thermodynamic limit $N\to\infty$, we obtain
\begin{equation}
C_{rs}
= 2 \sum_{k=1}^{\infty} \left( t_r^{k} - t_s^{k} \right)^2 .
\end{equation}
Expand the square and write
\begin{equation}
C_{rs}
= 2 \left[
     \sum_{k=1}^{\infty} t_r^{2k}
   + \sum_{k=1}^{\infty} t_s^{2k}
   - 2 \sum_{k=1}^{\infty} (t_r t_s)^{k}
   \right].
\end{equation}

Each series is geometric and convergent for $|t_r|,|t_s|<1$:
\begin{equation}
\sum_{k=1}^{\infty} t_r^{2k} = \frac{t_r^{2}}{1 - t_r^{2}}, 
\qquad
\sum_{k=1}^{\infty} t_s^{2k} = \frac{t_s^{2}}{1 - t_s^{2}},
\qquad
\sum_{k=1}^{\infty} (t_r t_s)^k = \frac{t_r t_s}{1 - t_r t_s}.
\end{equation}

Therefore the thermodynamic-limit expression for $C_{rs}$ is
\begin{equation}
C_{rs}
= 2\left[
     \frac{\tanh^2(\beta_r)}{1 - \tanh^2(\beta_r)}
   + \frac{\tanh^2(\beta_s)}{1 - \tanh^2(\beta_s)}
   - \frac{2\,\tanh(\beta_r)\,\tanh(\beta_s)}{1 - \tanh(\beta_r)\tanh(\beta_s)}
   \right].
\end{equation}

\subsection{Large $N$}
\begin{figure}
    \centering
    \includegraphics[width=0.32\linewidth]{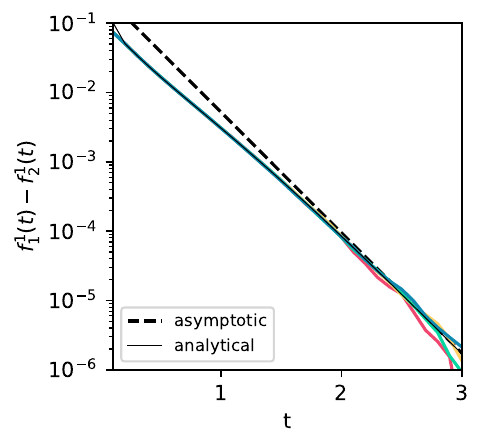}
    \includegraphics[width=0.32\linewidth]{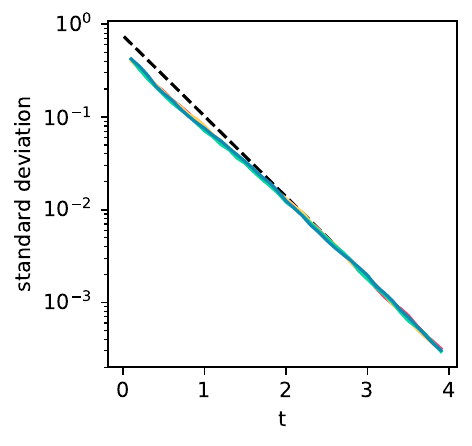}
    \includegraphics[width=0.32\linewidth]{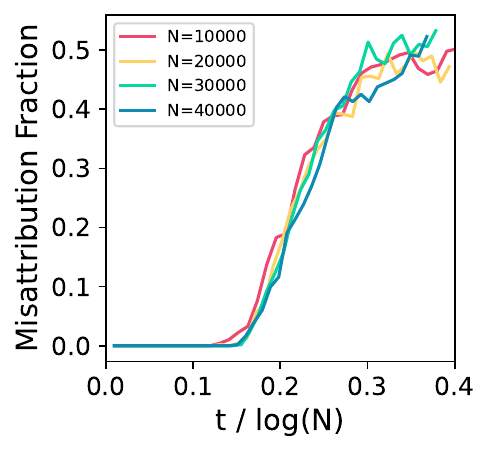}
    \caption{Scaling analysis for a mixture of 1D Ising models with inverse temperature $\beta_1=0.5$ and $\beta_2=1.0$. }
    \label{fig:largeN}
\end{figure}

For a mixture of 1D Ising models, we know how to compute analytically the left hand side of Eq.~\eqref{eq:speciation_crit}, but we do not have an analytical counterpart for the right hand side.
In section \ref{sec:ising} we described how to estimate speciation times approximating the right-hand side with its asymptotic value, obtained from a Gaussian approximation for large times, derived in Eq.~\eqref{eq:asympt_var}
\begin{equation}
    |f_{rr}(t_{rs})-f_{rs}(t_{rs})| = K \sqrt{\frac{C_{rs}}{2N}}e^{-2t_{rs}}.
\end{equation} 
This approximation holds when speciation times are large enough so that the Gaussian approximation is valid, namely for large enough $N$.
Since U-turn experiments are very expensive for large $N$, in Fig.~\ref{fig:free-entropy-forward} we limited ourselves to $N$ up to $3200$. For this reason, we resorted to the experimental value of the right hand side of Eq.~\eqref{eq:speciation_crit} to provide an accurate estimate of speciation times.
In this section, we report results for larger $N$ values that account only for the forward process. Nonetheless, they prove that for large times the asymptotic expression for free entropy difference average and variance are correct. In the left panel of Fig.~\ref{fig:largeN} we compare the average free entropy difference for a 2-Ising mixture with inverse temperatures $\beta_1=0.5$ and $\beta_2=1.0$ obtained numerically with the analytical value and with the asymptotic scaling. In the central panel, we do the same for the right hand side of Eq.~\eqref{eq:speciation_crit}, only without the analytical value that we did not compute analytically. 
Finally, in the right panel of Fig.~\ref{fig:largeN}, we verify that the scaling of the misattribution fraction coincides with the one of speciation, since rescaling time by $\log N$ all curves collapse. 

\subsection{Exact Score}\label{app:ising-exact-score}
The exact score function can be obtained from
\begin{equation}
   \mathcal{S}(x;t)=-\frac{x-e^{-t}\langle s \rangle_x}{\Delta_t} 
\end{equation}
where the average of $s$ under the tilted measure reads
\begin{equation}
    \langle s_i \rangle_x= \int ds \ s_i P(s \mid x)=\int ds \ s_i \frac{P(s,x_t)}{P(x_t)}
\end{equation}
In the Ising mixture case, one has
\begin{align}
    P(s, x_t)& =P_{0}(s)\frac{e^{-\frac{(x-se^{-t})^{2}}{2\Delta_{t}}}}{\sqrt{2\pi\Delta_{t}}^{N}} =\frac{1}{\sqrt{2\pi\Delta_{t}}^{N}} e^{-\frac{||x_t-s e^{-t}||^2}{2\Delta t}} \left(\sum_r w_r \frac{ e^{ \beta_r \sum_i s_i \sigma_{i+1}}}{Z_r}\right)
\end{align}
with $Z_r=2(2\cosh{\beta_r})^{N-1}$, leading to
\begin{equation}
    \langle s_i \rangle_x = \int ds \ s_i \ P(s \mid x)= \frac{\int ds \ s_i \ e^{\sum_i \frac{e^{-t}}{\Delta t} x_i^t s_i} \sum_r w_r \frac{e^{\beta_r \sum_i s_i \sigma_{i+1}}}{Z_r}}{\int ds \ e^{\sum_i \frac{e^{-t}}{\Delta t} x_i^t s_i} \sum_r w_r \frac{ e^{\beta_r \sum_i s_i \sigma_{i+1}}}{Z_r}}.
\end{equation}
Both the trace in the denominator and the average in the numerator can be computed through the transfer matrix method, for each value of $x$. For every $\beta_{r}$, the terms in the denominator are standard partition functions 
\begin{align}
B(\beta) =\int ds\ \frac{e^{\beta\sum_{i}\sigma_{i}\sigma_{i+1}+\frac{e^{-t}}{\Delta_{t}}\sum_{i}x_{i}\sigma_{i}}}{Z(\beta)} = \mathrm{Tr}\prod_{i=1}^{N}\frac{T_{i}^{\beta}}{z_{\beta}}
\end{align}
where 
\[
T_{i}^{\beta}=\left(\begin{array}{cc}
e^{\beta+\frac{e^{-t}}{\Delta_{t}}x_{i}} & e^{-\beta-\frac{e^{-t}}{\Delta_{t}}x_{i}}\\
e^{-\beta+\frac{e^{-t}}{\Delta_{t}}x_{i}} & e^{\beta-\frac{e^{-t}}{\Delta_{t}}x_{i}}
\end{array}\right)
\]
and $z_{\beta}=e^{\beta}+e^{-\beta}$. Instead, terms in the numerator will have the form
\begin{align}
A(\beta)_{j} =\int ds\ s_{j}\frac{e^{\beta\sum_{i}\sigma_{i}\sigma_{i+1}+\frac{e^{-t}}{\Delta_{t}}\sum_{i}x_{i}\sigma_{i}}}{Z(\beta)} =\sum_{s_{j}}s_{j}\mathrm{Tr}\left(\left(\prod_{i=j}^{N}\frac{T_{i}^{\beta}}{z_{\beta}}\right)\left(\prod_{i=1}^{j-1}\frac{T_{i}^{\beta}}{z_{\beta}}\right)\right).
\end{align}
Finally
\begin{equation}
\langle s\rangle_{x}=\frac{\sum_{r}w_{r}A(\beta_{r})}{\sum_{r}w_{r}B(\beta_{r})}.
\end{equation}

\end{document}